# A Hybrid NN/HMM modeling technique for online Arabic handwriting recognition

Najiba Tagougui, Houcine Boubaker, Monji Kherallah, Adel M. ALIMI
*REGIM-LAB: Research Groups on Intelligent Machines Laboratory, University of Sfax, National School of Engineers (ENIS), BP 1173, 3018, Sfax, Tunisia*
{najiba.tagougui, houcine-boubaker, monji.kherallah, adel.alimi}@ieee.org

**Abstract**

*In this work we propose a hybrid NN/HMM model for online Arabic handwriting recognition. The proposed system is based on Hidden Markov Models (HMMs) and Multi Layer Perceptron Neural Networks (MLPNNs). The input signal is segmented to continuous strokes called segments based on the Beta-Elliptical strategy by inspecting the extremum points of the curvilinear velocity profile. A neural network trained with segment level contextual information is used to extract class character probabilities. The output of this network is decoded by HMMs to provide character level recognition. In evaluations on the ADAB database, we achieved 96.4% character recognition accuracy that is statistically significantly important in comparison with character recognition accuracies obtained from state-of-the-art online Arabic systems.*

## 1. Introduction

Reproducing human ability in computer applications is still challenging despite the huge advances occurred in the field of artificial intelligence applications. The handwriting recognition domain with his two branches offline and online has been of interest for the last four decades. Actual results are very promising and according to the literature the recognition accuracy rates reported by many systems are over 90%. For the Arabic script and due to its writing characteristics which can be explained in more details in [1], (curliness, overlapping characters, presence of ligatures, diacritics use …), the state of the art is less in advance and it is far to be claimed as solved problem. Multiple approaches were developed and Hidden Markov Models (HMMs) are one of the most successful methods used as in [2] [3] [4] [5]. They are statistical models which have been found extremely efficient in the field of automatic speech recognition. This success has motivated recent attempts to implement HMMs in character recognition whether on-line or off-line. Their use in handwriting recognition can be explained by their capability to segment and recognize a handwritten script which can be very complex when using explicit methods. A variety of neural networks architectures have been used also [6] [7] [8]. They were used basically for isolated character recognition. This is principally due to the pre segmented data needed for the training [9]. Thus, a hybrid method combining the two types of the above classifiers seems to be promising to take advantages of large and non-linear context modeling via neural networks while profiting from the Markovian sequence modeling [10] [22]. Previous attempts exist in the literature and they are very successful for both online and offline handwriting recognition [11] [12] [13] [22].

In our case, the neural network is used to obtain the observation probabilities for HMMs. In the coming sections, we are going to detail the different stages of the system and to discuss the challenges of each stage. Finally, we will report our testing results based on the ADAB-database [14] [20].

## 2. System overview

In this section we present our proposed system architecture (see Figure 1) including the preprocessing and normalization steps. Then the extracted features are described according to the beta-elliptic modeling technique. The segmentation principle used is depicted and finally, we detailed the recognition system we used based on MLPNNs and discrete HMMs.

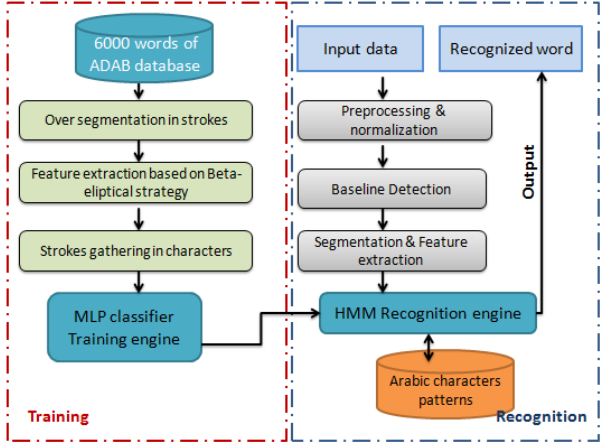

**Figure 1**. Proposed system overview

## 2.1. Preprocessing

The preprocessing step is essential in any handwriting recognition process to eliminate the noise generated when acquiring the signal on the digital devices and to handle the various speeds and shapes of writing by normalizing the handwriting dimensions. First, the vertical dimension of the script lines is adjusted to obtain a normalized size script. The following steps summarize the algorithm

1. The value *m* is computed:
   $m = \max(max\_x - min\_x, max\_y - min\_y)$;
2. For each *x, y* of the trajectory script, we change the initial values using these formula
   $x\_norm = 128 * ((x - min\_x)/m)$;

   $y\_norm = 128 * ((y - min\_y)/m)$;

   End for

Where 128 is a threshold value fixed after several experimental tests.

Then, a Chebyshev second type low pass filter at a cutoff frequency $f_{cut}$ of 12Hz and a radius of filtering window R = 8 is applied on the normalized trajectory to eliminate the noise introduced by temporal and spatial sampling (see Figure 2).

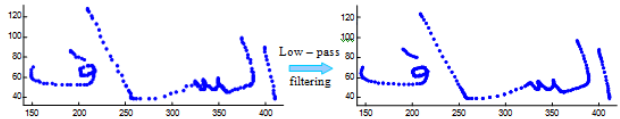

**Figure 2**. Low - pass filtering for noise elimination and trajectory smoothing

## 2.2. Feature extraction

The performance of the developed recognition system depends heavily on how well the different writing styles are modeled. Unlike the Latin script, the Arabic language has a large set of characters depending on their positions within a word [1]. Thus, recognizing a given character is a more difficult task. To extract the feature vectors from the normalized script, the enhanced Beta-Elliptical strategy [16] is used. The purpose is to decompose the signal into segments. Each segment is defined as a continuous handwriting stroke between two extremities points representing pen-up or pen-down moments [16]. The set of extracted features can be divided into two classes. The first class consists of dynamic features extracted for each point considering the velocity profile with respect to time by inspecting the extremum points representing local maximums and minimums of the curvilinear velocity signal of handwriting and its inflexion points. The second class takes static features representing handwriting trajectory. Each segment trajectory is modeled by arcs of ellipse as explained in [16]. The resulting feature vector is composed of 10 parameters: 6 dynamic features modeling the velocity profile (temporal information) and 4 static features modeling the trajectory aspect (geometric and spatial information).

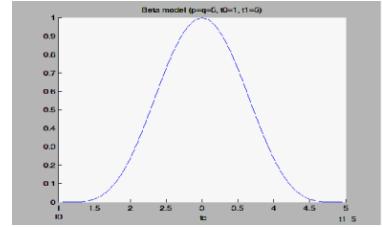

**Figure 3**. The Beta model

**2.2.1. Dynamic features.** According to the Beta approach, the generation of a handwritten velocity model is the algebraic result of adding the Beta velocity profiles of the successive strokes [16].

$$V(t) = \sum_{i=1}^{n} K.\beta(t, q, p, t_0, t_1) \quad (1)$$

$$\beta(t, q, p, t_0, t_1) = \begin{cases} \left(\dfrac{t-t_0}{t_c-t_0}\right)^p \left(\dfrac{t_1-t}{t_1-t_c}\right)^q & \text{if } t \in [t_0, t_1] \\ 0 & \text{if not} \end{cases} \quad (2)$$

- $\Delta t$ : t1-t0 time features describing the time interval delimiting the beginning and the end of the continuous stroke (the segment)
- *Tc*: the time corresponding to an inflection point in

the velocity profile
- *P*: it is parameter independent of the velocity profile determining the Beta function shape.
- *K*: the amplitude of the function Beta
- *Vi*: the initial energy at the beginning of the interval Δt which dissipates over time.
- *Vf*: the developed energy to conduct the continuous stroke until the end of the interval Δt.

**2.2.2. Static features**. A continuous stroke segment is modeled by two arcs of ellipse. This results in the following parameters:
- *a1, b1* and $\theta 1$ for the first arc of ellipse
- *a2, b2* and $\theta 2$ for the second arc of ellipse

Since $\theta 1 = \theta 2$ and a1 = a2 (the two angles represent the inclination of the tangent to the path at the point corresponding to tc), so that the trajectory modeling features may be limited to: *a1, b1, b2* and $\theta 1$.

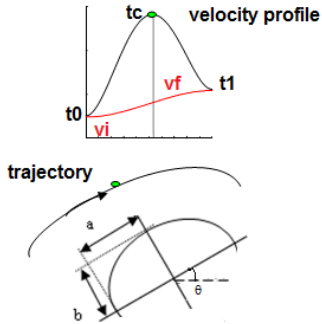

**Figure 4**. The velocity profile –trajectory correspondence

**2.2.3. Baseline detection.** Baseline detection is a step principally used for delay strokes detection or character segmentation and features extraction. In our case, we adopt the algorithm proposed by Boubaker et al. [21]. This virtual baseline is detected by following two different stages: the first one extracts sets of aligned points candidates to carry the baseline. The second stage executes a topologic evaluation on the candidates' sets to correct the detection result.

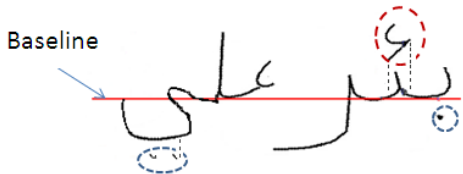

**Figure 5**. Detection of the virtual baseline and delayed strokes projection

**2.2.4. Delayed strokes handling.** The Arabic script has the specificity of using delayed strokes and dots in addition to the main body words leading to more complexity in the online recognition process. They are generally detected and ignored as in [17] or added in the post processing step as in [4]. In our system, we used almost the same idea as in [2] to handle them. The major difference is that in our algorithm, we first detect the virtual baseline to distinguish the word main body from the delayed strokes. Finally we vertically project the first point (corresponding to $t_0$ in the velocity profile) and the last point (corresponding to $t_1$ in the velocity profile) of the delayed stroke segment to the nearest non delayed stroke segment as shown in Figure 5.

In sum, the final feature vector is composed of six dynamic parameters, four static parameters and eleven parameters for baseline detection to handle delayed strokes yielding in a 21-dimensional feature vector.

$$\vec{x_t} = (\underbrace{x1..x6}_{\text{dynamic features}}, \underbrace{x7..x10}_{\text{static features}}, \underbrace{x11..x21}_{\text{baseline features}})$$

## 3. Segmentation

String or word recognition is generally more difficult than isolated character recognition due to the difficulty of character segmentation. In fact, characters cannot be reliably segmented before they are recognized. Therefore, the hard task in the segmentation process is how to validate the right segmentation point according to predefined character patterns. Moreover, many segmentation techniques are involved and they were discussed by Abuzaraida et al. [23].

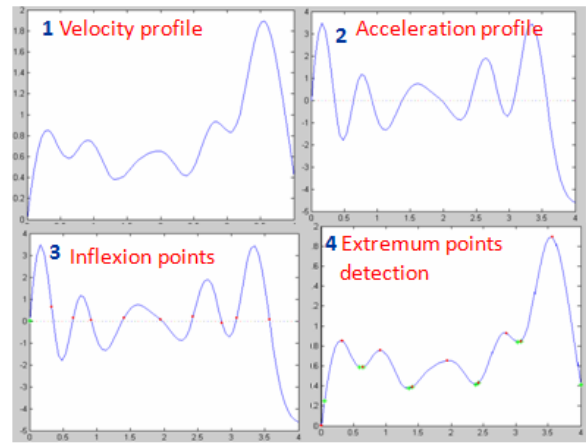

**Figure 6**. Extremum points detection process

In our approach, the handwriting trajectory is segmented in continuous strokes called segments [16]. Each stroke is defined as a continuous line between pen down and pen up movement by inspecting the local extremum points (maximum and minimum) in the curvilinear velocity signal. A pen lift separates two adjacent strokes. Each primitive segment (set of strokes) is assumed to be a character. Hence a word will not be split to characters but to segments. A strokes gathering mechanism is then needed to extract the different characters composing the segmented word. Examples of strokes gathering results are presented in Table1.

**Table1.** Strokes gathering results for characters "ب", "س" and "ح"

| Character | Position | \multicolumn{6}{c}{Strokes number} |
|---|---|---|---|---|---|---|---|
| | | 2 | 3 | 4 | 5 | 6 | 7 |
| ب | Be | ☒ | ☒ | | | | |
| | Mi | | ☒ | | | | |
| | En | | | ☒ | | | |
| | Iso | | ☒ | | | | |
| س | Be | | | | ☒ | ☒ | |
| | Mi | | | | | ☒ | ☒ |
| | En | | | | | ☒ | ☒ |
| | Iso | | | | | ☒ | |
| ح | Be | ☒ | ☒ | | | | |
| | Mi | | ☒ | ☒ | | | |
| | En | | | ☒ | ☒ | | |
| | Iso | | ☒ | ☒ | | | |

## 4. Recognition engine

In this section we briefly summarize HMMs and MLP NNs. We then introduce our hybrid approach for online Arabic handwriting recognition.

### 4.1. Hidden Markov Models

Hidden Markov models (HMMs) are able to segment and to recognize at the same time which is the major reason for their use for automatic handwriting recognition systems. The idea of applying HMMs to handwriting recognition was originally motivated by their success in speech recognition [9].

For the analysis of sequential data, the use of HMMs as statistical models can be considered the state-of-the-art [26]. They are, generally, based on the assumption of a statistical model for the generation of the data to be analyzed. The purpose is to find the sequence $\hat{w}$ that maximizes the posterior probability $P(\mathbf{w}|\mathbf{X})$ of the symbol sequence given the data X.

$$\widehat{W} = \arg\max P(w|X) = \arg\max \frac{P(w)P(X|w)}{P(X)}$$
$$= \arg\max P(w)P(X|w)$$

Given an input text sequence $X$, the recognizer should find the sequence $W$ of words which maximizes the probability $P(w) P(X | w)$. The term $P(w)$ is the prior probability of the word which is estimated by the language model. $P(X | w)$ is the observation likelihood estimated using the language model.

### 4.2. Mutlilayer Perceptron

Among artificial neural networks, Multilayer Perceptron (MLPs) are one of the most popular network. They are trained by the backpropagation algorithm [24]. It has been shown in various papers that MLPs with a single hidden layer are universal classifiers, in the sense that they can approximate decision surfaces of arbitrary complexity, provided the number of neurons in the hidden layer is large enough [25].

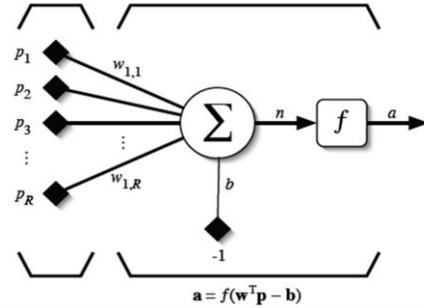

**Figure 7.** Artificial neural network model

Where
- $p = [p_1 p_2 \ldots p_R]^T$ is the input vector
- $w = [w_{1,1}\ w_{1,2}\ \ldots w_{1,R}]^T$ is the weights vector.
- $n = \sum_{j=1}^{R} w_{1,j}\ p_j - b$
  $= w_{1,1}p_1 + w_{1,2}p_2 + \cdots + w_{1,R}p_R - b$
  $= w^T p^T - b$

Each neuron has one output $a$ which is the result of the transfer function $f$ of the weighted sum of its inputs.

### 4.3. Hybrid MLP/HMM classifier

Neural networks can be considered statistical classifiers under certain conditions, by supplying output of a posteriori probabilities. Thus, it is interesting to combine the respective capacities of the HMM and the MLP for a new efficient recognition system inspired by the two formalisms. The major question is how to combine the two different classifiers especially that such a system isn't simple to implement because of the number of parameters to adjust and the large amount of training data necessary to ensure the global model. In this section, we show how our hybrid system is designed.

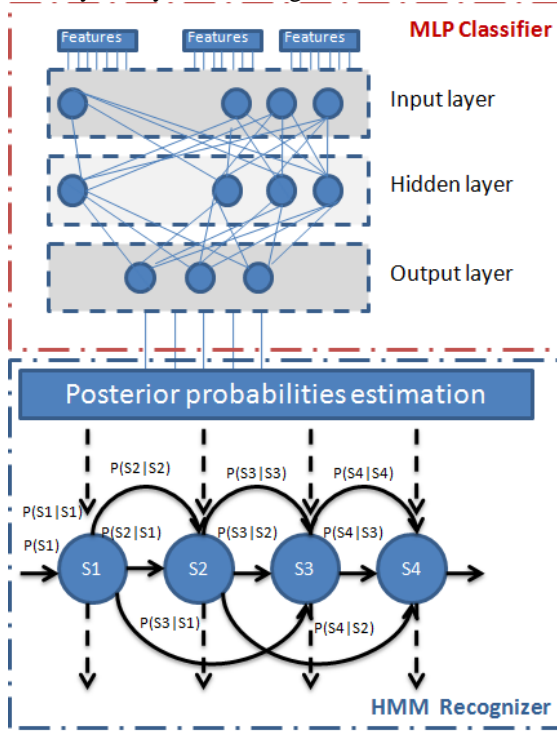

**Figure 8**. The proposed hybrid MLPNN/HMM recognition structure

The proposed recognition system is MLPNNs feature extracted based to convert class probability estimates to posterior probabilistic feature vector suitable for subsequent HMMs recognizer. Our MLPNN system is composed of OCONs (One Class One Neural Network). Each class of a handwritten character corresponds to one OCON. The architecture of OCON is presented in Figure 4. Each neural network was trained by the standard back propagation algorithm with training parameters (the rate: $l = 0.01$; the momentum factor: $a = 25$; and the iterative number for training: epochs = 4000) as in [18]. Because of the multi-variability of the handwriting, every character has not got the same strokes number. This number is variable and takes a value from 2 to 7.

For the Arabic handwriting script, each one of the 28 characters of the alphabet can have up to 4 different shapes depending on his position within a word (at the beginning, in the middle, at the end or isolated) [1]. We note also the use of ligatures. Besides the previous classes, the ADAB database used to validate our work contains some digits and ligatures. As a result, a total of 120 characters modeled by a multi-state left-to-right discrete HMM are used. Each character was modeled by a 4 state HMM. In fact, the number of states per model cannot be computed. It is fixed after experimental tests (see Table 2). The first state and the final one are non-emitting states and are used to provide transitions from one character model to the other character model. Each state has self loops and transition to adjacent states with just one skip. The Viterbi algorithm is used to train the HMM proposed system with the maximum likelihood criterion.

The principal idea behind the MLPNN/HMM hybrid approach as illustrated in Figure 8, is to estimate the output probability density function of each state of the used HMM by the output nodes of the MLP classifier which received segment stroke features as input. These input vectors are preprocessed to finally estimate the posteriori probability deciding whether the input vector belongs to the desired character class. The MLPNN's output, weighted by the priori probability of each class, forms the probability density function used for every state of the HMM.

The MLP posterior probabilities p(S|X) are divided by the prior state probabilities p(S) in order to approximate the observation probabilities of our HMM.

$P(X|S) \approx \frac{P(S|X)}{P(S)}$ as described in [22] where

$-\log p(X|S) = -\log p(S|X) + \alpha \log p(S)$    (3)

With α being a priori scaling value leading to no improvements in the used log-linear function. Hence, the obtained posteriori probabilities are tied together. After training the MLPNNs, the Viterbi Algorithm is used to train the discrete HMMs. Those discrete HMMs are used to model the input signal by using vector quantization to transform density vectors into discrete symbols. This task needs a codebook which defines a set of clusters of the feature space. The codebook size used is 256 whose value is also fixed after several experimental tests (see Table 2).

**Table 2**– Average Recognition rates using different codebook sizes and number of states per model

| Code book size | Number of states per model | | | | |
|---|---|---|---|---|---|
| | 2 | 3 | 4 | 5 | 6 |
| 126 | 72.32% | 80.25% | 82.50% | 83.51% | 78.56% |
| 256 | 80.56% | 87.89% | 91.23% | 90.27% | 88.96% |
| 512 | 82.78% | 86.58% | 87.14% | 86.19% | 85.80% |

| | | | | |
|---|---|---|---|---|
| Sum | 21575 | 33164 | 174690 | 166 |

For the training module, we choose to use segmented characters instead of using a database of isolated characters to better address problems of characters concatenation in an input word. Therefore we segmented 6000 words chosen randomly from the first three sets of ADAB. We obtained 378950 segment strokes that we have injected to the MLPNN to assemble different Arabic characters skeletons which will be used later by the HMM recognizer. The strokes gathering mechanism was supervised and rather some steps were performed manually to ensure good segments classification.

In the evaluation phase, we applied the system on sets 4, 5 and 6 of ADAB. To evaluate the performance of the proposed system, we compared to standard MLP system and to discrete HMM and continuous HMM recognition system. The training data was the same for both systems.

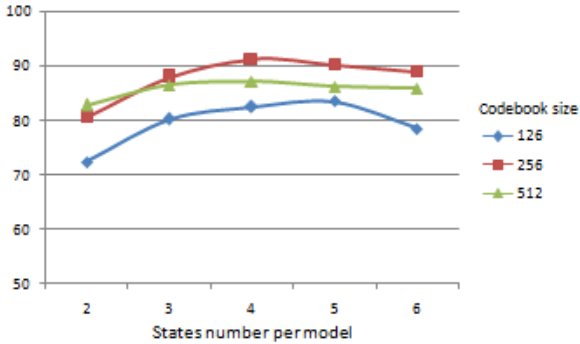

**Figure 9**. Average Recognition rates using different codebook sizes and number of states per model

## 5. Experimental results and discussion

### 5.1. Experimental setting

The experiments were conducted using the ADAB database which includes 937 different labels of online Tunisian town's names. Details of the database are presented in Table3.

**Table 3**- Different ADAB datasets

| Set | Files | Sub-words | Characters | Writers |
|---|---|---|---|---|
| 1 | 5037 | 7670 | 40500 | 56 |
| 2 | 5090 | 7891 | 41515 | 37 |
| 3 | 5031 | 7730 | 40544 | 39 |
| 4 | 4417 | 6786 | 35832 | 25 |
| 5 | 1000 | 1551 | 8189 | 6 |
| 6 | 1000 | 1536 | 8110 | 3 |

### 5.2. Results and discussion

The achieved recognition results are presented in the following table 4. First, it can be seen, that the discrete system outperforms the continuous HMMs. This agrees with previous works in the literature as in [11] and [22]. We notice also that the results presented by the MLP classifier are significantly lower. They are principally due to the lack of data during the learning phase.

Compared to the standard MLP system, to discrete HMM system and to continuous HMM system, the proposed system presents a better recognition rate for the three test sets.

**Table 4** - Results of the different systems on the test sets 4, 5 and 6 of the ADAB database

| Systems | Set 4 | | | Set 5 | | | Set 6 | | |
|---|---|---|---|---|---|---|---|---|---|
| | Top1 | Top5 | Top10 | Top1 | Top5 | Top10 | Top1 | Top5 | Top10 |
| Basic MLP system | 71.14% | 71.14% | 71.14% | 69.52% | 69.52% | 69.52% | 70.52% | 70.52% | 70.52% |
| Continuous HMM system | 90.15% | 91.89% | 91.89% | 89.97% | 91.23% | 91.23% | 90.71% | 91.87% | 91.87% |
| Discrete HMM system | 91.26 % | 93.25% | 93.25% | 91.62% | 92.78% | 92.78% | 90.36% | 92.78% | 92.78% |
| MLP/HMM system | 96.45% | 97.08% | 97.58% | 96.12% | 97.51% | 97.51% | 95.20% | 96.86% | 97.27% |
| Best ICDAR 2011 system | 98.02% | 98.98 % | 98.98 % | 98.18% | 99.18% | 99.18% | 98.45% | 98.97 % | 98.97% |

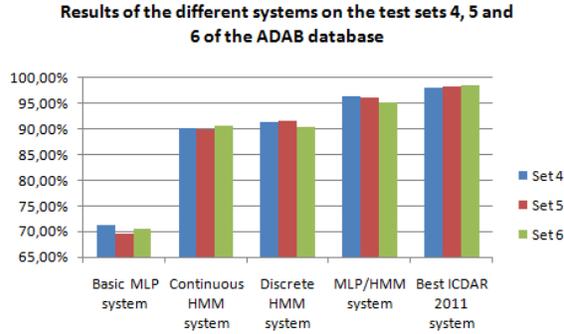

**Figure 10**. Results of the different systems on sets 4, 5 and 6

The results of our hybrid MLP/HMM system are slightly lower than those achieved by the first system that has competed for the online Arabic handwriting recognition competition in ICDAR 2011 [20]. Besides, our system can be used for applications dealing with open lexicon because it is character based.

## 6. Conclusion

In this work we proposed a hybrid MLPNN/HMM system for Arabic handwritten script recognition where the output probability density functions of the discrete HMM's states are approximated with the MLP neural net. The purpose was to take advantages of the discriminative power of the neural network while profiting from the Markovian sequence modeling. To avoid a higher decoding complexity mainly due to the use of the both classifiers, MLPNN were just used on the training part of the system.

In an experimental section we showed that there is significant gain in recognition rate compared to a baseline discrete HMM system where the absolute recognition rate was improved by 3.51% with the proposed hybrid approach. Future work consists of improving the actual proposed system following two main directions: feature extraction and solid recognition engine. Additional features may enhance the current feature vector like introducing offline parameters widely used by recent approaches. Improved HMM models can also be built by adjusting the optimal number of states per model.

## Acknowledgment

The authors acknowledge the financial support of this work by grants from the General Direction of Scientific Research and Technological Renovation (DGRST), Tunisia, under the ARUB program 01/UR/11/02.